\documentclass[a4,12pt]{article}
\font\sevenrm=cmr7

\def\@{@}
\catcode`\@=13
\def@#1{\csname Pol#1\endcsname}
\def\smfont{\sevenrm}
\def\Polcedil#1#2{\leavevmode\hbox{\setbox0\hbox{\smfont `}%
#1\hbox to 0pt{\kern-#2\wd0\lower.9\ht0\box0\hss}}}

\usepackage[english]{babel}
\usepackage{graphicx}
\usepackage[T1]{fontenc}
\usepackage{times}
\usepackage{amsfonts}
\usepackage{amssymb}
\usepackage{amsthm}
\usepackage{titlesec}
\usepackage{anysize}
\usepackage{url}

\usepackage{graphicx}

\newtheorem{Defi}{Definition}

\newtheorem{Pro}{Proposition}

\begin{document}


\title{\bf A Note on Systematic Conflict Generation in CA-EN-type
Causal Structures}

\author{Antoni Lig@eza\thanks{On leave from: 
Institute of Automatics AGH, al. Mickiewicza 30,
30-059 Krak@ow, Poland; e-mail: {\tt ali\@ia.agh.edu.pl}. The  
author's stay in LAAS du CNRS  was supported by a  MENESRIP-DAEIF
scholarship No.: 174755 K through CIES. } \\
{\small LAAS du  CNRS, 
7, Av. du Colonel Roche,  31077 Toulouse Cedex, France} \\
{\small E-mail: {\tt ligeza\@laas.laas.fr}}, \\
}

\date{}

\maketitle


\begin{abstract}
This  paper  is aimed  at  providing a  very first, more  ``global'',
systematic point of view with  respect to possible conflict generation
in CA-EN-like causal structures.  For  simplicity, only the  outermost
level of graphs is taken into account.  Localization of the ``conflict
area '', diagnostic preferences,   and bases for  systematic  conflict
generation are considered. A notion of {\em Potential Conflict
Structure} ({\em PCS}) constituting a basic tool for identification of
possible conflicts is proposed and its use is discussed.
\end{abstract}

\section{Introduction}

Diagnostic reasoning is an activity oriented towards detection of
faulty behaviour and its explanation, i.e. isolation of faulty
components responsible for the observed misbehaviour of the analyzed
system. Model-based diagnosis is based on explicit system model
applied for diagnostic  inference. A widely accepted approach consists
in {\em consistency-based} reasoning where the analysis is aimed at
regaining consistency of the predicted model output with current
observations by retracting some of the assumptions about correct
behaviour of certain components. The sets of elements suspected to
contain at least one faulty component are identified by detecting
inconsistency between the observed and predicted behaviour. Such sets,
called {\em conflict sets} are basic products for generating diagnoses.

A complete diagnostic procedure following the consistency-based
approach should cover: 

\begin{itemize}
\item detection and localization of misbehaviour,
\item restriction of the search area (hierarchical fault diagnosis),
\item systematic conflict generation, taking into account that:
\begin{itemize}
\item {\em all} conflicts should be found, but
\item only {\em minimal} conflicts should be generated,
\end{itemize}
\item diagnoses generation,
\item verification and repair.
\end{itemize}

This paper is mostly concerned with systematic conflict
generation. The problem of conflict generation appears to be 
one  of the most important  problems in automated diagnosis of dynamic
systems based  on domain model  of correct system behaviour.
 
Conflict
sets \cite{Reiter} (or {\em conflicts}, 
for short) are the sets of components of the
system such that under the assumed model and  observed output not {\em
all} the components of   any conflict set  cannot  be claimed to  work
correctly. Such  sets  of ``suspected''   elements are used  then  for
potential diagnoses generation in the form of hitting sets for all the
conflicts (i.e. a diagnosis  is any  set having nonempty  intersection
with  any conflict set, and  build from the  elements of conflict sets
only).   This kind of diagnostic  approach is based on Reiter's theory
\cite{Reiter} of diagnosis  from first principles, and  DeKleer's work
on  diagnostic  systems  \cite{DeKleer}.    In application to  dynamic
systems the theory   describing system behaviour  is  constituted by a
causal qualitative model  of correct system behaviour  in the form  of
CA-EN causal graphs incorporating qualitative calculus \cite{Bousson}. 

When considering the problem of conflict set generation, the following
simplifying assumptions will be considered to hold: 

\begin{itemize}
\item the causal  qualitative model is complete  in the sense that the
behaviour of all variables of  interest can be effectively  calculated
(simulated) under  the  assumption  of  correct behaviour  of   system
components, 
\item the possibly  observed incorrect behaviour of certain  variables
is due  to one or  more faults  of  components  only; no  misbehaviour
caused   by incorrect design or implementation, closed-loops feedback
effects,  wrong control  actions  (input variables), external noise,
impreciseness of the model or measurements are taken into account, 
\item the potential faults can be caused  only by elements ``assigned''
to  influence   relations represented  by edges    of  the graph; for
simplicity one can assume that one influence is represented  by 
 one ``identifiable component'' assigned to it, 
\item quasi-static faults are considered only, i.e. faults that can be
observed for  some time period, causing steady-state-like misbehaviour
observed during some time interval; no temporal misbehaviour is
considered here, i.e. faults are assumed to be of permanent nature, 
\item the structure of the subgraph of interest does not change during
its analysis,
\item the considered graph has no loops,
\item no time (dynamics) is taken into account,
\item all the influences are ``calculable'', i.e. the equations
describing the signal propagation are solvable both in the forward and
in the backward direction,
\item all the misbehaving variables can be detected. 
\end{itemize}

Further,  no knowledge   about  potential misbehaviour  modes  of  the
components  will be used. The  assumed goal is  to find, possibly all,
explanations, i.e. faults, responsible for the observed misbehaviour. 

\section{Graphical notation}

For the sake  of representing graphically  various features concerning
analyzed  cases a simple  extension of  the causal  graphs symbolics is
proposed.  Throughout  the paper the  following extension of the basic
notation of \cite{Bousson} will be used: 

\begin{itemize}
\item $P$, $Q$, $R$ -- input/control  variables, 
\item $U$, $V$, $W$ -- intermediate variables, 
\item $X$, $Y$, $Z$ -- output variables; $X$ is also used as any
variable (without making precise its position), 
\item $[U]$, $[\; ]$ -- a variable not measurable, 
\item $X$, $\underline{X}$, $\bullet$ -- a  measured variable, 
\item $X~$, $X^{\ast}$, $\ast$ -- a variable observed to misbehave, 
\item  $X{+}$,  $X^{+}$,  $X{-}$ , $X^{-}$     -- extended notions  of
misbehaviour, i.e. providing the information that the value is too big
or too small, respectively (reserved for future use). 
\end{itemize}

As usually, an  arrow ($\longrightarrow$) means causality. Families of
variables are to  be  denoted with  boldface  characters.   e.g {\bf X}   =
$\{X_{1}, X_{2},\ldots,X_{k}  \}$.  Influences (equations) are denoted
by $I$,   e.g.   $P \stackrel{I}{\longrightarrow}  X$ means   that $P$
influences  $X$ through $I$; a component  responsible  for the correct
work of $I$ is to be  denoted by $c$.   Faulty components or influences
will be also  denoted by $c^{\ast}$  and $c{\ast}$, respectively.  For
simplicity, assuming that one  component  $c$ is responsible for   the
correct behaviour of influence $I$, we can interchange components with
influences and vice versa. 

In  case of dynamic  equations a  ``time-flattening'' procedure may be
applied. This means that a differential  equation can be replaced with
an appropriate  set of algebraic  equations describing the relationship
among the  variable values in the  subsequent time  instants, as it is
done for numerical solution of differential equations.

\section{Causal graph}

The class of  considered causal graphs is  quite a general  one.  By a
{\em  causal  graph} we understand  a  set of  variables (taking either
numerical or symbolic values) and  represented by the graph nodes, and
a  set of {\em causal influences}  defined with appropriate equations,
and represented by the arcs of the causal graph.  Thus any causal graph
is assumed to be a  structure of the form   {\bf G} = $({\bf X},  {\bf
\Psi})$, where {\bf X}  is  the set of  all  the variables and   ${\bf
\Psi}$ is the set  of influences/equations allowing for calculation of
certain  non-input variables on the  base  of the  input  ones. It  is
assumed that the equations   defined by ${\bf  \Psi}$ are  forward and
backward calculable, i.e. having all input variables for one influence
the  output  can be  calculated, and  having  the  all but  one  input
variable   values and the value   of  the output  variable, the single
undefined input  variable can be calculated.  It is  no matter here if
the calculation are analytical or numerical ones.

\section{Basic problem formulation}

A reasonable  assumption is   to start  the  diagnostic procedure   at
discovering that  at least one   of the observed (i.e. measurable  and
measured)  variables  misbehaves. This can be   done  basically in two
ways: 

\begin{itemize}
\item the detection can  be a {\em model-based}  one, i.e. the value of
the variable is predicted  (by simulation, under assumption of correct
work of all  the components) and  compared with the observed value. In
case significant discrepancy  is observed, the  variable is classified
as misbehaving  or non O.K.  This method  is in  fact applied in CA-EN
\cite{Bousson};  some   further details  considered  there  cover the
problem of  minimal time period  of the discrepancy being observed (to
avoid   false alarms   caused  by fluctuations,   etc.),  and problems
following  from qualitative character of the  values of the considered
variables. 
\item the detection can be  based on {\em expected behaviour} approach
\cite{Ligezab,Report}, i.e. the definition of {\em expected normal behaviour}
or {\em  expected failure behaviour}   can be stated  explicitly.  The
definitions can be based both on the analysis of  the model and on the
expert domain knowledge and former experience. The advantage of the
latter approach is that the detection can be quicker, as it does not
require simulation with use of the system model.
\end{itemize}

Moreover,  any combination     of the above   two   approaches  can be
applied.  In case of  complex systems with  qualitative models and big
degree  of uncertainty and  vagueness in the  domain knowledge, such a
combination may be necessary in order to reasonably  cover most of the
real failures and  avoid   covering the false ones.   Further, general
``integrity constraint'', e.g. in the  form of logical formulae can be
provided   to  describe   consistent and    inconsistent   patterns of
variable/values combinations \cite{dx96}.

The starting point for diagnostic  procedure consists in determining a
nonempty  set    of   misbehaving   variables    ${\bf    X}^{\ast}  =
\{X_{1}^{\ast},X_{2}^{\ast},\ldots,X_{k}^{\ast}  \}$.  We assume  that
during  the operation   of   diagnostic procedure  this    set remains
unchanged, i.e. quasi-static  faults are to   be diagnosed.  Our  goal
here is   to determine  possibly all  minimal   conflict sets  for the
observed set ${\bf X}^{\ast}$. This conflict sets may be used then for
diagnoses generation. 

It seems reasonable to distinguish the  following stages to be carried
out in course of a systematic conflict generation procedure: 

\begin{enumerate}
\item   {\bf Domain restriction:} 
 Restriction (possibly maximal)  of the   initial   graph to a
subgraph containing only  the variables and components ``involved'' in
the creation of observed misbehaviour; however, certain {\em boundary}
variables may  also be useful,  e.g.   to eliminate certain  suspected
components   and/or  to   further  structure   the  potential conflict
sets.  Note that in a  more  complex system several independent faults
may  occur at a   time  in ``geographically separated''  areas  of the
system;  it seems   reasonable to  perform   diagnostic reasoning then
independently for any such area,
\item {\bf Strategy selection:} 
Establishing a strategy  for conflict generation, e.g.   ``hot''
starting points,  order of generation, restrictions,   etc. One of the
key  issues   there is that   the generation  of   conflicts should be
efficient both with respect to time of generation  and with respect to
their ``parameters'' (conflict sets  should be as precise as possible,
i.e. minimal),
\item {\bf Efficient conflict generation:}
    Systematic conflict   generation,     usually from from  the
``smallest'' to   the  ``biggest''  ones with  deleting   non-minimal
conflicts and efficient elimination of potential conflicts sets which are
not real conflicts. This point is crucial for the diagnostic
    efficiency -- if not all the conflicts are generated, some faulty
    elements may be missing in final diagnoses; if conflicts are not
    minimal, too many diagnoses are likely to be obtained.

\end{enumerate}

The conflict  generation stage  can    be interleaved with   diagnoses
generation. A post-analysis of conflicts after generation stage may be
useful as   well; elimination of certain  conflicts  -- if possible --
leads to smaller diagnoses,  while considering minimal conflicts leads
to   smaller  number of potential diagnoses.     In  order to establish
efficient   strategy both   {\em    preferences}  and  current    {\em
limitations}  should be  taken  into  account. Further,  expert domain
knowledge and    heuristics may  be  useful  (e.g.   in the    form of
pre-schedules or plans for ordering conflict generation).

Below we consider the three above stages as separate problems.

\section{Graph restriction}

Let us consider the problem  of restricting the  area of focus in order
to minimize the domain for potential conflict calculation. This can be
done by restricting the initial graph {\bf G} to  a subgraph, say {\bf
G'},  sufficient for the diagnostic  task. Intuitively, the goal is to
rule out most of the  variables and influences not  taking part in the
formation of  the observed misbehaviour.  We discuss below some of the
most straightforward possibilities.

Consider the most  simple and   intuitive restriction consisting    in
limiting the area of interest to elements  from which there is a signal
flow to the  misbehaving  variables (as in \cite{Bousson}); 
 in other   words,  the idea   of
causality is   applied to elimination of items   not having the causal
influence on the observed misbehaviour. 
Let $ANT(X)$ denote the subgraph composed of components (influences)
such that through any $c$ of $ANT(X)$ there is a directed path to $X$.
Similarly,   let   $DESC(X)$ denote  the   subgraph  composed  of  all
components (influences)  such that there is a   directed path from $X$
through any  $c \in DESC(X)$. Similar notation  can be applied to sets
of  variables. Only the   elements   involved  in $ANT(X)$ can    have
influence on the behaviour of $X$.

It would  seem  natural to limit   the area  of  interest to $ANT({\bf
X}^{\ast})$ but the simplest example concerning back-propagation shows
that in some cases   it may be not   sufficient (see Fig.1). 


\begin{figure}[!htb]
 \centering
  \includegraphics[width=0.95\textwidth]{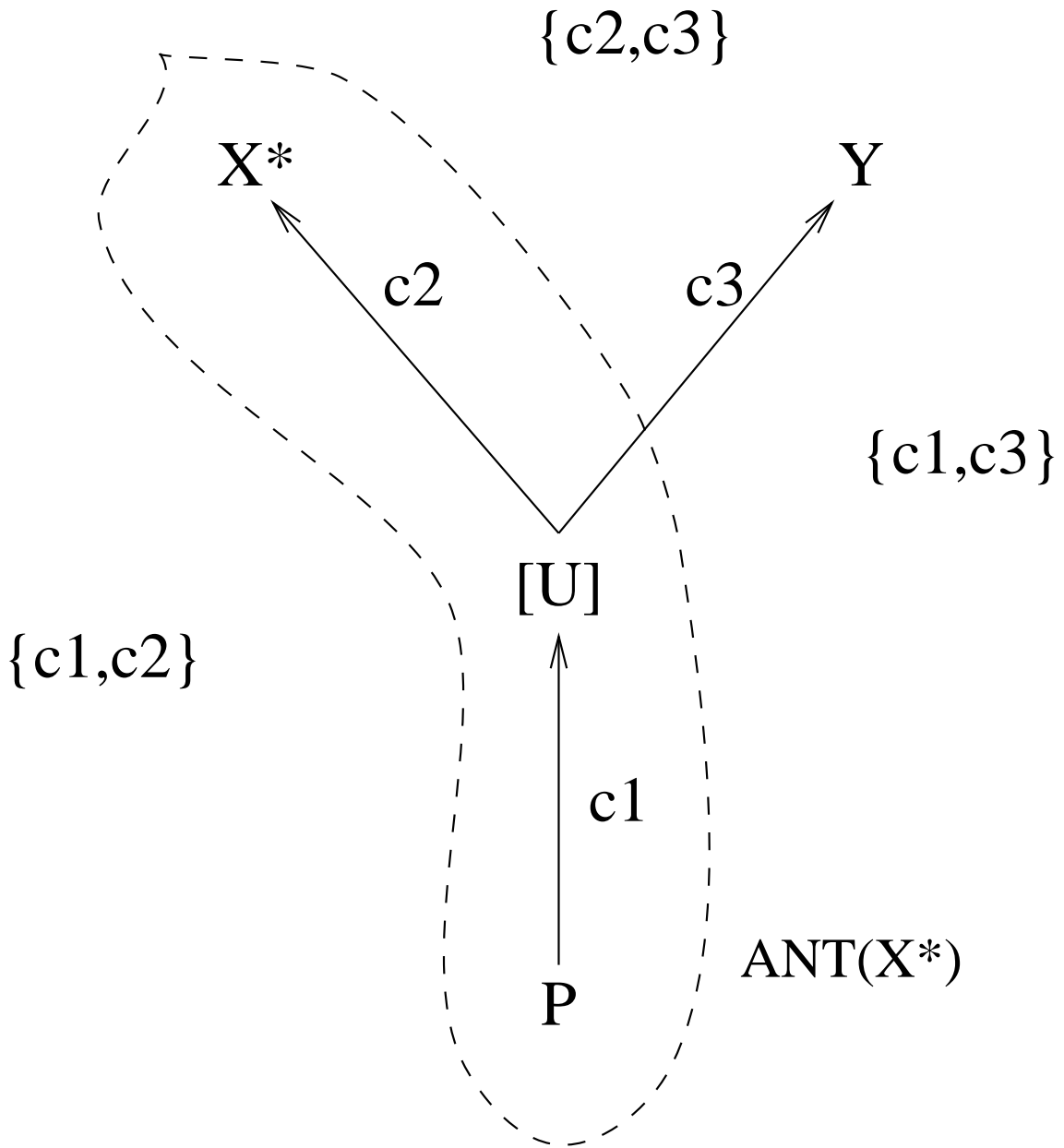}
  \caption{Example -- restriction  of the domain to $ANT({\bf X}^{\ast})$.}
  \label{fig1}
\end{figure}


Note that,
in such a case at least two problems following from missing of some auxiliary
information occurs:  
\begin{itemize}
\item lack of intermediate point checking,
\item compensation phenomenon for multiple faults in line.
\end{itemize}
 In case
$c_{1}$ is faulty $c_{3}$ may  compensate for its misbehaviour so that
$Y$  behaves correctly.    In   such a  case   even   if $c_{3}\not\in
ANT(X^{\ast})$, it should certainly be taken  into account. The use of
it may be twofold: if $\{c_{2},c_{3} \}$ is a conflict set (apart from
$\{c_{1},   c_{2}\}$),  then probably  $c_{2}$  is  faulty; however the
explanation    that $c_{1}$  and  $c_{3}$    are  faulty and   $c_{3}$
compensates  for the fault  of $c_{1}$  at $Y$  is also  possible (the
compensation  phenomenon takes place).  In  case $\{c_{2},c_{3} \}$ is
not a  conflict  set, then most probably  $c_{1}$  is  the only faulty
element there. Thus the part with  $c_{3}$ provides new discrimination 
information.      We  shall refer  to     this  type of structures  as
``fork-like''. 

The next approximation may be to limit the area of analysis to the set
defined as  $DESC(ANT({\bf   X}^{\ast}))$  and it seems  to   be quite
reasonable\footnote{This, in fact, is the core of the approach
 equivalent to  assuming that back-propagation is not
combined with propagation forward.}.  However, again, extension of the
former recent example shows   that  in certain cases   this  heuristic
simplification   may   lead    to  incomplete  conflict     generation
possibilities (see Fig.   2).

\begin{figure}[!htb]
 \centering
  \includegraphics[width=0.95\textwidth]{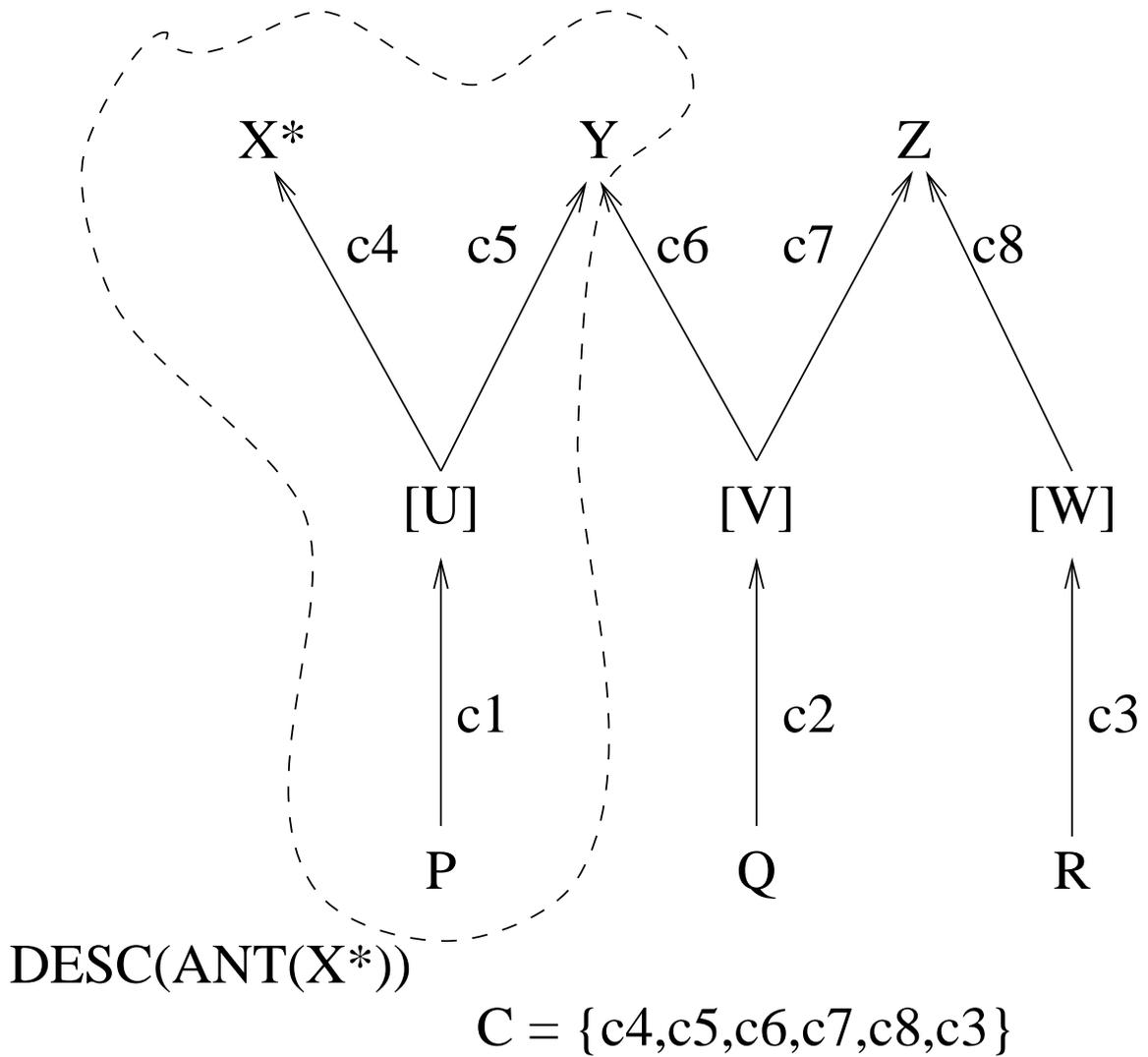}
  \caption{Example -- Restriction of the domain to $DESC(ANT({\bf X}^{\ast}))$.}
  \label{fig2}
\end{figure}


 One of   the possible conflict sets  is
$\{c_{3},c_{8},c_{7},c_{6},c_{5},c_{4}  \}$, and in order to calculate
it  one has to  consider a complex combination of ascending/descending
elements.  Of course this example  may be extended horizontally on any
arbitrary number of variables; we shall refer to it as ``side-wave''
or ``side-effect'' example. 

The next obvious possibility is an arbitrary combination of the form 

\[
DESC(ANT( \ldots DESC(ANT({\bf X}^{\ast})) \ldots )) 
\]

or similar; the main problem is where  to stop.  Of course, an obvious
solution  is to  cover  all the  connected  graph, i.e.  break the
procedure of subgraph growing on the  ``natural boundaries''. Below a
more  restrictive   proposal,  still  satisfying   the requirement  of
generating all possible conflicts is outlined. 

Let  us  consider an  arbitrary  subgraph of the initial  causal graph
containing  at least one  unmeasured variable.  By {\em extending} the
graph  we shall understand  adding subsequent  links (with assigned  to
them variables). Extending on a  certain path (paths)   to or from  an
unmeasured  variable leads to a  {\em closure} if all the ``boundary''
variables are measured ones, and the values of the unmeasured variables
incorporated in the subgraph 
can be  calculated (either by for-   or by back-propagation) from
the   boundary   variables.  The smallest    set  of measured boundary
variables defining a closure for the variables of the initial graph on
all the paths to or  from it ``cuts out''  the subgraph of
interest. It is  to
be denoted by  $CLO({\bf X})$ for the  initial set  of variables being
{\bf X}.  In  other words,  the $CLO({\bf  X})$  is  a minimal  subgraph
covering  ${\bf  X}$,  ``cut off''  from  the  basic   causal graph at
measured variables only (and including all of them).  Below we propose
a formal definition following the above intuitions. 

\begin{Defi}
Let {\bf   X} be  an arbitrary   set of  variables  (both measured and
unmeasurable ones). The closure of {\bf X}, to be denoted as $CLO({\bf
X})$ is a subgraph of the causal
graph satisfying the following conditions:

\begin{itemize}
\item $CLO({\bf X})$ incorporates all the variables of {\bf X},
\item all the    input and output    variables of $CLO({\bf  X})$  are
measured ones, 
\item all the input and output variables are the O.K. ones,
\item $CLO({\bf X})$ is minimal with respect to set inclusion of the set 
of subgraph nodes. 
\end{itemize}
\end{Defi}

The meaning of input and output variables is straightforward. An input
variable is one from which the signal is directed inside the structure
and taking it value from outside the  structure. An output variable is
one taking its value  from inside the  graph and, if some links  point
from   this variable they must  all   go outside the   closure. Thus a
variable to which several links point  can be a boundary variable only
if all the   links are pointing inside  or  outside the  closure.  For
example,  on  Fig.  2  variable  $\underline{Y}$  is  not  a  boundary
variable for the  structure incorporating  elements $c_{1}$, $c_{4}$,
and $c_{5}$. 

\begin{figure}[!htb]
 \centering
  \includegraphics[width=0.95\textwidth]{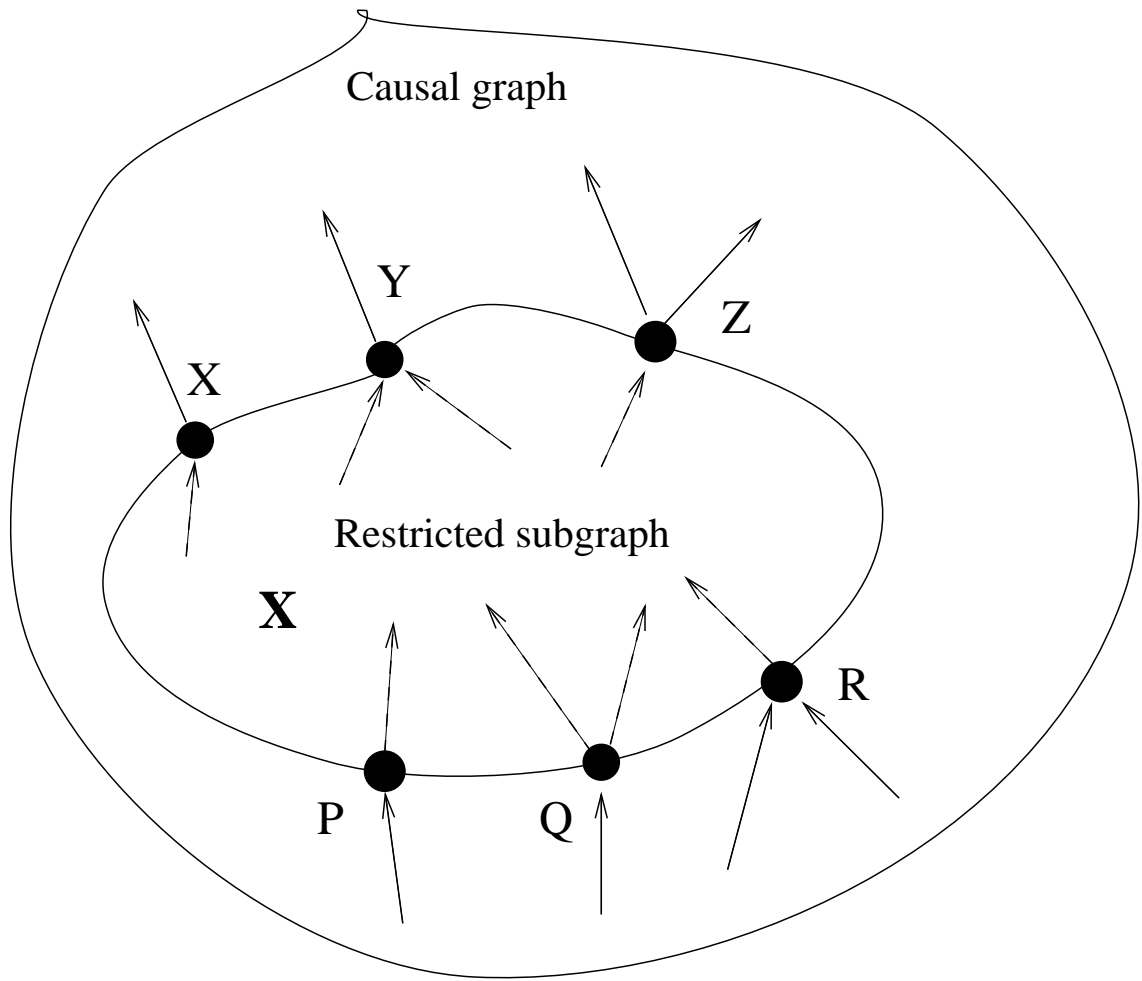}
  \caption{An information closure.}
  \label{fig3}
\end{figure}

For  intuition, the  construction  of  the  $CLO({\bf X})$  cuts out a
specific subgraph from the initial graph, but one can cut only through
measured   and   correct   variables  and  not   through   links  (see
Fig. \ref{fig3}).

 This means  that one should try  to isolate the subgraph with respect
to information coming in or out of the subgraph;  one tries to isolate
it from information theory  point  of view.   Of course, the  selected
subgraph  should   be  the  smallest one,    covering  the misbehaving
variables.  Further, the variables through which  we cut should behave
correctly - the   goal is to isolate  misbehaving  part of  the graph.
This   is  also like  growing the   misbehaving  area  until correctly
behaving measured variables are reached on all paths going outside. 

The input and  output variables  will be referred  to  as {\em boundary
variables}; in fact, they constitute a kind of  a frontier such that all
the   information  coming    in or out      goes  through the  boundary
variables. Intuitively, if    they behave O.K.    no information about
misbehaviour inside $CLO({\bf X})$ can  go out from the structure  and
thus be manifested  outside; similarly, no information  about possible
faults outside can go inside the structure and thus be a cause for the
observed misbehaviour of the incorporated variables.

Note that the above operation separates in fact  a subgraph closed from
the point of view of information theory; assuming the ``Markow-like ''
character of  the causal  graphs, no  information  can be  transferred
inside  or outside  the $CLO$-sured  subgraph in  a  way different than
through the boundary variables.  Thus, no information from the outside
can have influence  on the diagnostic process,  provided that the goal
is to explain the misbehaviour of the variables inside the $CLO$-sure.

Now  the natural consequence   of the above   idea is to restrict  the
subgraph for conflict generation to be $CLO({\bf X}^{\ast})$ (see Fig.
3  for and intuitive  idea).  Moreover, if several separated subgraphs
consisting closures for the  misbehaving variables can be constructed,
the diagnostic process can   be  performed for any identified   sub-area
independently from  each other. Roughly  speaking,  in such  a case the
closure would constitute  ``islands''of isolated  misbehaviour on  the
area of the initial causal graph. 

The idea of the  algorithm for constructing $CLO({\bf X}^{\ast})$ may
be as  follows.  First construct   a $CLO(\{  {X}^{\ast}  \})$ for any
variable $X^{\ast} \in {\bf X}^{\ast}$.  This can be done by following
any path  to  and  from $X^{\ast}$ until   a   measurable and  correct
variable is spotted. Then any closures of single variables sets having
some elements in common  should be composed into  connected subgraphs.
The process may result with one or more connected subgraphs. 

Note that the isolation  process can be done  for both misbehaving and
O.K. variables,  i.e.  one can perform  a  construction like the above
closure for a set  of correct variables. In  this case, such a closure
should be excluded   from the diagnostic   process  (maybe without  the
boundary variables).

As it can be  seen,  the  above choice is  quite  a natural one, and   it
follows also  from   the approach  to conflict  calculation  presented
later.

\section{Strategy for conflict calculation}

In order to consider the problems of strategy of conflicts generation,
one should first  answer  the questions  concerning preferences  among
diagnoses  to  be generated and  sets  of diagnoses  viewed as ``final
solutions ''. 

Recall   that any diagnosis $D$ is   just a set  of   components, $D =
\{c_{1}, c_{2},  \ldots, c_{m}\}$  such  that assuming  them faulty is
sufficient for restoring the consistency of the domain theory with the
observations.   The sets  of   diagnoses are  denoted  as ${\bf   D} =
\{D_{1}, D_{2}, \ldots, D_{n} \}$.. 

Considering preferences among diagnoses we must take into account the
risk of basing our diagnostic procedure on incomplete set of conflicts,
i.e.   a case when not  {\em  all}  the conflicts  are calculated. 
Intuitively,    the diagnoses calculated   in   such  a case  will  be
``incomplete'' or ``partial''.  This  observation is supported by the
following proposition. 

\begin{Pro}
Let ${\bf C}_{i}$ denote  sets of conflict sets,  and ${\bf  D}_{i}$ 
sets of  diagnoses calculable from ${\bf C}_{i}$,  $i = 1,2$.  Then, if
${\bf C}_{1} \subseteq {\bf C}_{2}$, then also: 
\begin{enumerate}
\item $||{\bf D}_{1}|| \leq ||{\bf D}_{2}|| $, and 
\item $\forall D_{1} \in {\bf D}_{1} \; \exists D_{2} \in {\bf D}_{2}$
such that $D_{2} \supseteq D_{1}$. 
\end{enumerate}
\end{Pro}

According  to the  above proposition,  in  case of  more conflict sets
accessible  (i.e. known; in  our case the  problem is  to discover the
conflicts  since they  are ``hidden''   in  the graph  structure), the
diagnoses are possibly more precise\footnote{At  this stage we  assume
to  prefer as  precise  (complete) diagnoses as  possible.  By a more
precise diagnosis we    understand here one  explaining more    of the
observed  failures;  contrary  to   precise (complete) diagnoses,  the
imprecise  (incomplete)    ones explain  only   part  of  the observed
misbehaviour.  This may   seem contrary  to  the  usual preference  of
minimal  diagnoses.  Our standpoint is,   however, following from the
risk that  not all the  conflicts are likely to  be generated.  For an
established  set of conflicts  preferring minimal diagnoses  is still an
obvious choice.}.  Another  observation is that simultaneously, having
more conflicts we can expect generation of more diagnoses, and this is
what  we would  like to  avoid. As the   diagnoses  are only potential
explanations  of   the observed  misbehaviour,   they require  further
verification,  thus  the  number   of diagnoses  generated  should  be
relatively small. The  intuition is that  generation  of the conflicts
should be  limited  to  the minimal ones  (or  at  least as  small  as
possible).  The following proposition  is  related to  the  problem of
limiting the number of possible diagnoses.

\begin{Pro}
Let ${\bf C}_{i}$ denote sets of conflict sets, and ${\bf D}_{i}$ sets
of  diagnoses  calculable from  ${\bf C}_{i}$,  $i  =  1,2$.  Further,
assume that ${\bf  C}_{1}$ and ${\bf C}_{2}$  have the same  number of
elements. Then, if for any $C_{1} \in {\bf C}_{1}$ there exists $C_{2}
\in {\bf C}_{2}$ such that ${C}_{1}  \subseteq {C}_{2}$, then there is
also $||{\bf D}_{1}|| \leq ||{\bf D}_{2}|| $. 
\end{Pro}

The above proposition allows for comparison of two sets of conflicts
with the same number of elements, but different ``degree of
preciseness''; the more precise conflicts are better, they lead to
generating less possible diagnoses.

Finally, let us consider the problem of adding new conflict sets to an
existing set of conflicts. If in the already found conflict set there
is one smaller than the one to be added, then adding the new one is
not necessary; either no new diagnoses will be generated or the
diagnoses will not be minimal. The following proposition states it
more precisely.

\begin{Pro}
Let ${\bf C}_{i}$ denote sets of conflict sets, and ${\bf D}_{i}$ sets
of diagnoses  calculable from  ${\bf C}_{i}$, $i   = 1,2$.   Let ${\bf
C}_{2}$ contain all the conflicts of ${\bf  C}_{1}$ and some other not
minimal ones, i.e. ${\bf C}_{2} = {\bf C}_{1}\cup {\bf C}_{1}'$, where
for any $C_{1}'\in {\bf  C}_{1}'$ there exists $C_{1}\in {\bf  C}_{1}$
such that $C_{1}\subseteq C_{1}'$.  Then for any diagnosis  $D_{2}\in
{\bf D}_{2}$  there  is a diagnosis  $D_{1}\in {\bf  D}_{1}$ such that
$D_{1}\subseteq D_{2}$.  Further,  there is also $||{\bf D}_{1}|| \leq
||{\bf D}_{2}|| $. 
\end{Pro}

This proposition justifies the intuition that non-minimal conflicts are
useless for diagnostic efficiency -- adding  non minimal conflicts not
only may  lead  to more  diagnoses,   but also  to generation of   non
minimal ones as well. 

The above considerations   seems to justify the  following assumptions
concerning the strategy of conflict generation: 

\begin{enumerate}
\item Conflicts  should be generated in  a systematic way, so that all
the necessary conflicts   are obtained; if  a  conflict  set is
missing, there is a risk of generating partial (incomplete) diagnoses. 
\item Conflicts  should be  generated from  $i = 1$  towards $i  = k$,
where $i$ is the number of components in a conflict set and $k$ is the
maximal number  of components in  the  analyzed subgraph; this assures
that more precise conflicts are generated first. 
\item All   conflicts  comprising $i$  elements   should be calculated
before  ones comprising  $i  +  1$ elements;  a   conflict which  is a
superset  of  some   previously  generated  conflict  is    not to  be
considered. 
\item The procedure can be stopped when either no new conflicts can be
generated, or  for any new conflict to  be generated a subset conflict
has already been generated (minimality requirement). 
\end{enumerate}

Note that the  following further auxiliary  rules may be  proposed for
enhancing he diagnostic process: 

\begin{itemize}
\item the conflict generation procedure may  be stopped for some number
of conflicts  generated   arbitrarily;  this may   be  the  case  when
diagnoses containing only some limited number of faulty components are
probable, 
\item the conflict  generation procedure may also  be stopped for  some
$i$ arbitrarily; this  may be the  case when too complex conflicts are
too costly to calculate, etc., 
\item more  data  (measurements, tests) may  become  available cutting
down in a natural way the size of conflict sets, 
\item conflict generation may be interleaved with diagnoses generation
(see also \cite{Reiter});  some diagnosis  found  valid may  stop  the
process, 
\item  expert-designed  {\em schedules}   for  conflict generation
finding the most   probable conflicts ``around''  elements most likely
exhibiting faulty behaviour can be used to speed-up the diagnostic
procedure by turning  it into routine procedures, 
\item if accessible, the knowledge  about modes of faulty behaviour of
the components and  its influence on the  behaviour of variables can be
used for further selection. 
\end{itemize}

\section{An approach to systematic conflict generation}

The basic assumption here is that both propagation and back-propagation
can be regarded as  {\em mathematical constraints}, i.e.  they provide
some equations determining the   relations among variable  values.  An
equation with one unknown value  of a variable  defines this value; if
all the values are known, then such an equation provides a possibility
of conflict  generation -- the components  responsible for  holding of
this equation may  not be all working  correct if the equation is  not
satisfied by the observed variable values. 

In   case of back-propagation  there   can  be  some difficulties   with
``inverting'' the calculations  so as to  obtain the values for one of
the arguments.  However, from theoretical point  of view it seems that
in any case  one can solve  the equation numerically, and so inverting
it is not a    necessary procedure; of  course,   if applicable, it   can
contribute to computational efficiency. 

\begin{figure}[!htb]
 \centering
  \includegraphics[width=0.95\textwidth]{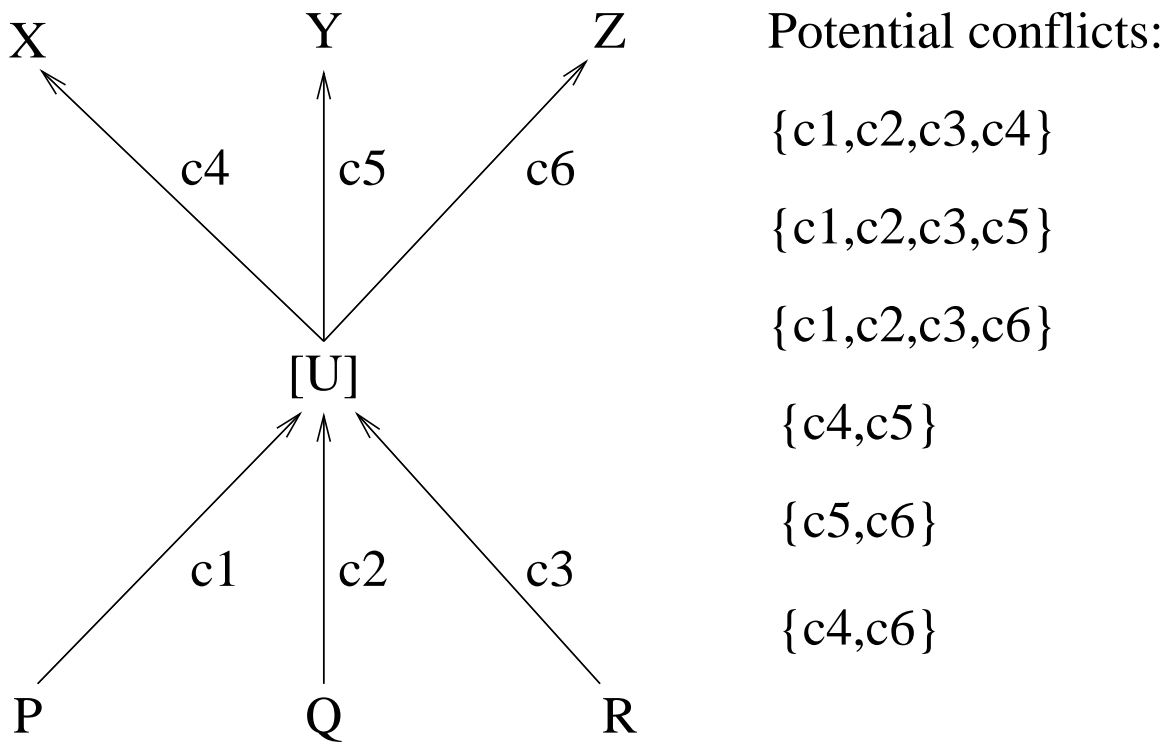}
  \caption{Example -- potential conflict structures selection.}
  \label{fig4}
\end{figure}

From purely mathematical point of view, in order to generate conflicts
one  must have more  equations  than variables; in  our  case they are
unmeasured variables.   Thus  if $n$  denotes the number  of equations
(both for  back- and for-propagation) defined  for some  subgraph, and
$m$   is  the number  of    unmeasured variables involved in the computation, then  the  condition
necessary  for potential conflict  generation  is  that $n \geq  m+1$.
Further,   for  any  such    substructure  there   exists a  potential
possibility of  generating  no more  than ${n  \choose m+1}$ possible 
conflicts. This can be  illustrated with Fig.\ref{fig4}, where  $n = 4$, $m = 1$,
and we have $4 \choose 2$ potential conflicts.


Usually, there   are  less conflicts,  since  not   all the structures
described by $n$  equations  and containing  $m$ variables allow   for
calculation of a ``full-size''  conflict; examples include chains with
pending unmeasured variables. Further, all the conflicts are only {\em
potential} --   if a conflict  is really  observed  or not  depends on
actual computations, and,    of course, on   the existence  of  faulty
elements  (it  is assumed that   no  conflicts  are generated  due  to
inadequate calculations or inadequate model). 

Taking into account the  above considerations and  in order to achieve
better efficiency of conflict generation, the following, 
two-stage, transparent procedure of conflict generation is proposed; 

\begin{itemize}
\item identification of {\em potential conflict structures}, i.e. sets
of  influences  assuring    {\em necessary}   conditions  for conflict
existence, and then 
\item verification  for any  such  a  structure  and selected  set  of
equations  if a conflict  exists; this is to  be done by an attempt at
``solving'' these equations. 
\end{itemize}

Splitting the procedure of  conflict generation into these  two stages
seems  to advantageous  for the  sake  of transparency and  systematic
conflict generation.  Moreover, identification of a conflict structure
is equivalent   to having the   knowledge  about it  components.   Thus
exploration of non-minimal  potential    conflicts can be    abandoned
without performing  the    real  calculations.     Moreover,   certain
heuristics  can be  applied    to  preselect the  potential   conflict
structures  for further investigation,  leaving  a large  part  of them
without performing costly mathematical calculations.

The key issue for  carrying on is to  introduce a definition of a {\em
Potential Conflict Structure}, shortly {\em PCS}.  This notion denotes
a   subgraph of the  causal  graphs,  for  which there  is a possibility
(always  potential)  of calculating    a conflict via    obtaining two
different values for  the  same  variable.   A $PCS$  comprising   $m$
unmeasured variables and leading to detection of potential conflict at
a  variable  $X$  will  be denoted   as  $PCS_{m}(X)$.  The  number of
unmeasured variables $m$ will  be referred to as  the {\em order} of a
conflict structure. Variable $X$ can be measured one or an unmeasured
one. 

Note that some most   interesting  are potential conflict   structures
having no unmeasured variables, i.e.   $PCS_{0}$ -- they are always of
the form  $P_{1}, P_{2},  \ldots, P_{j}  \stackrel{c}{\longrightarrow}
X$, where all the variables are measured; if such a structure provides
a real conflict, then the conflict consists of one  element $c$ and in
fact is a partial diagnosis.  In  other words, component $c$ is faulty
and must be  an element of any valid  diagnosis; further  the fault of
$c$ is a cause of the observed misbehaviour of $X$. Therefore conflict
structures of   zero  order  should always     be explored first   (if
existing). 

Now let us pass to potential conflict structures of larger size. First
we put forward the following definition. 

\begin{Defi}
A variable $X$ is \underline{well-defined} (defined, for short) iff: 
\begin{itemize}
\item either it is a measured variable, or 
\item its value can be calculated on the base of some other variables,
which are well-defined. 
\end{itemize}
If there are two or more, e.g. $k$ independent ways of calculating the
value of $X$, then $X$ is said to be $k$-defined. 
\end{Defi}

The  independent   ways of calculating  the    variable may consist in
measuring   the  value and  calculating  it   with  different sets  of
equations.  The calculation of a  well-defined variable can be done no
matter how -- forward propagation is as good  as backward one (at least
from   purely mathematical point  of   view).   Now we  can  define  a
potential conflict structure on $m$ unmeasured variables. 

\begin{Defi}
A \underline{\bf Potential Conflict Structure} for variable $X$ defined on
$m$ unmeasured  variables is  any subgraph  of  the causal graph, such
that:
\begin{itemize}
\item  it comprises exactly $m$  unmeasured variables (including $X$, if
unmeasured), 
\item  all    the variables   are  well-defined,   and   $X$ is
double-defined;
\item   for the   $PCS$   being  defined on  $m$   unmeasured
variables it is necessary that all the values of the $m$ variables are
necessary for making $X$ double-defined.
\end{itemize}
 Variable $X$
will be called the {\em head} of the PCS. 
\end{Defi}

Examples of $PCS_{m}$ for $m \in \{ 0,  1,2 \}$ are  shown on Fig.  5.

\begin{figure}[!htb]
 \centering
  \includegraphics[width=0.85\textwidth]{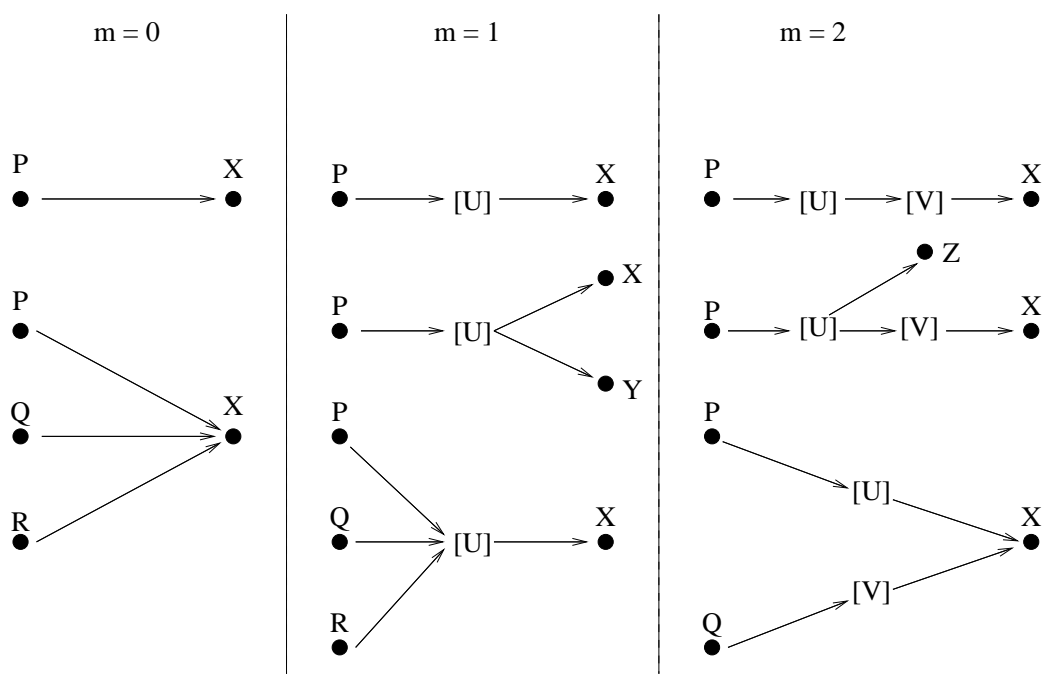}
  \caption{Examples -- potential conflict structures.}
  \label{fig5}
\end{figure}


Note that any of the graphs represent at least two potential conflict
structures, i.e. ones for different head variable;  in these cases the
$PCS$   constitute the same     graphs  and the  same  conflicts  will
eventually  be generated.  Thus for any  such $PCS$ the calculation is
to be performed only once,  and the selection of  the head variable is
to be done arbitrarily, e.g. with respect to making easier the problem
of equations solving. 

Potential conflict structures can take arbitrary  ``shape'' and it is,
in general,  difficult to say if   some structure is   a $PCS$ at the
first sight. The definition is in fact recursive, and so the algorithm
for  detection  of $PCS$  must  be.  But there  are  at least three 
typical  conflict    structures  with   some  nice    properties   and
interpretation.  They are: a ``chain'' of
calculable variables, a ``pyramid'' and  various types of ``forks  '',
where back-propagation plays important role. 

The algorithm for detecting a  $PCS$ on $m$  unmeasured variables is a
recursive one; the starting point is  a selected variable, the head of
the  structure; then, it    must  search for  exactly $m$   unmeasured
variables connected to the selected  one if it  is a measured variable
(a kind of path tracing), or $m-1$ unmeasured variables if the head is
an unmeasured  variable itself. When traversing the   graph a check if
all the unmeasured  variables are well defined  and if having them the
head  variable  is  double defined is    to  be performed.  The  basic
procedure should  be   repeated  for any  selected  variable   for $m$
changing from $0$ to the maximal number of unmeasured variables. 

Note that the  basic procedure  does not  guarantee  that the conflict
sets are generated  according to growing  number of  elements; this is
so, because  there can be  many  influences ``pointing  to'' a single
unmeasured variable and thus a potential conflict set may contain many
elements assigned to these influences.  However,  for any $PCS_{m}$ it
is  possible to  assign  a number  $j$  of elements  assigned  to the
influences necessary for conflict calculation; this can be done before
the calculation of  the appropriate conflict; there  is also always $j
\geq  m+1$.  Thus there   is  a  simple  way   of  defining a   second
characteristics of any $PCS$,  i.e. a  number  $j$ of elements of  the
conflict  set  to  be   possibly  determined; this  may  be   noted as
$PCS_{m}^{j}$.  And finally,  during calculation of conflicts  for the
same $m$ for any unmeasured variable, the  order of calculations can be
done with respect to the value of $j$.

\section{An outline of algorithmic approach}

To  summarize,  an outline  of  an  algorithm for systematic  conflict
generation can be as follows: 

\begin{enumerate}
\item Detect the set of misbehaving variables ${\bf X}^{\ast}$, 
\item  Define the restricted subgraph  being the object of analysis to
be $CLO({\bf X}^{\ast})$;  any other choice  (e.g.  a heuristic one) is
possible, but the completeness can be violated, 
\item For any measured variable $X \in CLO({\bf X}^{\ast})$ detect all
the conflicts calculable without the  use of the values of  unmeasured
variables; the calculation of conflicts can be ordered with respect to
the number of elements in conflict sets; this step refers to the
zero-order conflict generation, 
\item For any variable $X \in CLO({\bf X}^{\ast})$ detect sequentially
all $PCS_{m}^{j}$; the  order of  generation is from  $m  = 1$ to  the
number of variables in the subgraph of interest.  For any variable and
$m$ established order the $PCS_{m}^{j}$ according to increasing values
of $j$, 
\item  Repeated  $PCS$, i.e. ones  different only  with respect to the
head variable should   be abandoned (leaving exactly  one  of  them for
investigation); further, $PCS$ leading to non-minimal conflicts can be
abandoned before numerical investigation, 
\item Stop the procedure when there are no  more $PCS$ to be generated
(or earlier, according to   some  heuristics or when  an  appropriated
diagnosis is generated). 
\end{enumerate}

\section{An Example}

A simple test program for support of 
direct determining potential conflict structures was
implemented in {\sc Prolog}. The program is a meta-interpreter using
several simple recursive rules. It calculates the potential conflict
structures for a specified unmeasured variable.

\begin{figure}[!htb]
 \centering
  \includegraphics[width=0.45\textwidth]{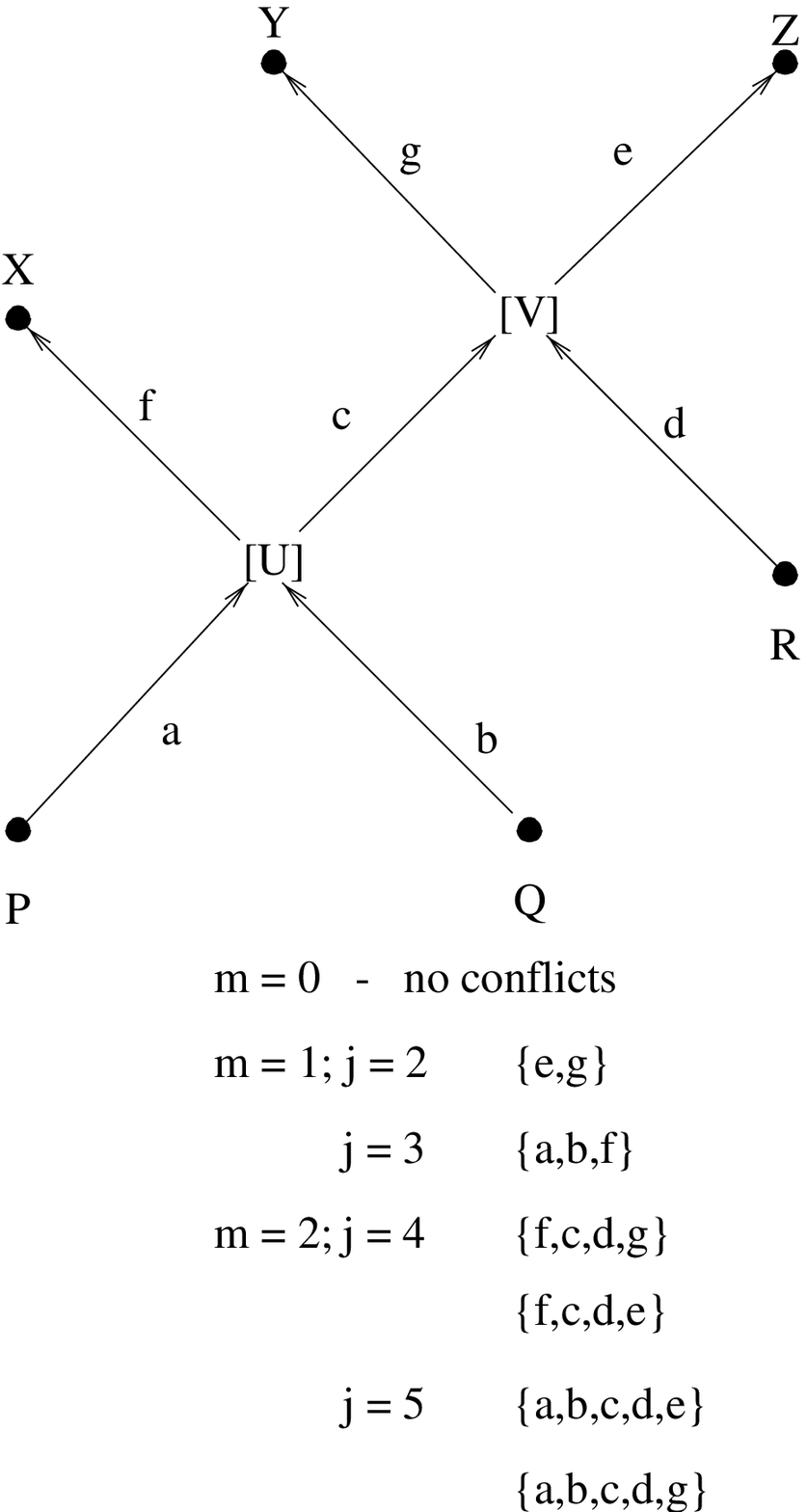}
  \caption{Example -- generation of PCS-s.}
  \label{fig7}
\end{figure}

An example subgraph $CLO$-sured   with measured variables is shown  in
Fig. \ref{fig7}; potential conflicts calculated with use of the program 
are listed there as well. 

During calculation of the conflicts the  following rules may be useful
in  order to  avoid  repeated  computations  and improve  the  overall
efficiency: 

\begin{itemize}
\item  {\em  head variables preselection:}  this   seems to be  a most
important one heuristic  rule for achieving reasonable  efficiency; the
candidate  variables for heads of  the $PCS$ should  be preselected. A
reasonable heuristic  may consists  in  selecting all  the misbehaving
variables and the unmeasured  ones (other variables, i.e. the measured
O.K. ones, will be incorporated in the  calculated $PCS$ or simply are
not  necessary; a strong  simplification  may consist in selecting the
non O.K.  variables as heads for $PCS$-s,
\item  {\em eliminating repetitions:}  the concept of $PCS$ allows for
identification of  conflict   set elements before calculation   of the
potential   conflict;  thus, whenever  a  $PCS$   identical to another
already generated appears, there is  no need to calculate the conflict
once more. This is important,  since starting from different variables
and extending  the $PCS$-s  it  seem unavoidable to  generate the same
$PCS$ several times. 
\item {\em  limiting the  size  of generated conflicts:}  again,  if a
$PCS$ has been generated such that its  elements constitute a superset
of a  conflict set which has already   been generated the  there is no
need to investigate this $PCS$, 
\item   {\em further decomposition:}  generation  of  conflicts can be
stopped at a boundary composed  of measured variables; this is similar
to considering the $CLO$-sure, but this time for a substructure of the
selected subgraph, 
\item  {\em  re-use  of  calculations:}   once calculated,  influences
(values of  the variables) can   be reused in calculation  of  several
conflicts; they need just to be stored, 
\item    {\em user-defined  scenarios:}    some  {\em schedules}   or
expert-defined  {\em  scenarios} providing  guidance  for calculating
conflicts in  specific cases defined by   the selection of misbehaving
variables  and type of  their misbehaviour  can be  used  to guide the
procedure of conflict generation, 
\item {\em  measurement introduction:} for  certain $PCS$  seeming too
large,  suggestions  of measurement   points can  be  done   so as  to
structure them down into manageable objects. 
\item {\em pre-elimination of potential faulty elements:} the observed
nature of faults may indicate that certain  types of faults are not to
be  taken into account; this  observation may lead to eliminate certain
components from further considerations.   Hence,  even if the  set  of
influences used  for  conflict  generation  are large,   the generated
conflicts may turn out to be quite small. 
\end{itemize}

\section{Closing remarks}

From the above considerations one can expect that the final efficiency
of conflict (not   diagnoses!)  generation is   bound to depend on   a
variety of factors. Further,  one can expect that efficient conflicts,
i.e.  ones  leading to a small  number of well-localized diagnoses can
be obtained only if the unmeasured variables are relatively sparse. In
case there are    only few measurements one  cannot   expect  that the
isolation of faulty components will be effective. 

With respect  to this problem, the  back-propagation seems to appear to
be a crucial issue -- roughly speaking, it may play  a role similar to
introducing measurements and thus contribute  to limiting the size  of
conflicts. 

Another aspect which seems well  worth investigating is the problem of
modes of faulty behaviour of the components and their influence on the
observed behaviour of variables;  an a priori knowledge about  possible
faults  of components can  be used to  model  the faulty behaviour and
thus to  select out certain  possible faults if the modeled behaviour
is not  observed.  Further,  certain  conflict sets can  be eliminated
during generation. 

But  the most  challenging issue   seems to   be  formalization of  an
approach  based on combination of direct search  for faulty components
(as in \cite{Diag96}) with procedures based on conflict generation;
Let us recall  that
the approach based on conflict generation according to Reiter's theory
\cite{Reiter}  is   justified only   under  several relatively  strong
assumptions; further, after generating   conflicts  a next stage   for
generating diagnoses is necessary.   The overall procedure seems to be
hardly   fitting on-line systems  requirements (not   to say about the
real-time ones). On the other hand, the idea of such a backward-search
procedure seems to
follow from the general  search principles and abductive reasoning: by
appropriate use  of  functional  element descriptions and  information
about  measured values one  should construct the hypotheses explaining
the observed behaviour.  In order to consider some more specific bases
for such an approach,  an ``axiomatization'' of the  domain seem to be
necessary.  Then the basic step    should consist in  generation of  a
search  space for  abductive diagnostic  reasoning,  providing a model
fitting the diagnostic purposes. 

\vspace{0.5cm}

\noindent
{\bf Acknowledgment:}  The
Author thanks Dr. Louise Trav\'{e}-Massuy\`{e}s for many comments and
 discussions helpful in improving this work.

\end{document}